\documentclass{article}
\usepackage{ijcai23}

\usepackage{times}
\usepackage{soul}
\usepackage{url}
\usepackage[hidelinks]{hyperref}
\usepackage[utf8]{inputenc}
\usepackage[small]{caption}
\usepackage{graphicx}
\usepackage{amsmath}
\usepackage{amsthm}
\usepackage{booktabs}
\usepackage{algorithm}
\usepackage{algorithmic}
\usepackage{xcolor}
\usepackage[switch]{lineno}
\usepackage{multicol}
\usepackage{multirow}
\usepackage{amssymb}

\newcommand{\mb}[1]{\mathbf{#1}}
\newcommand{\mc}[1]{\mathcal{#1}}

\newcommand{\lyx}[1]{{#1}}

\newcommand{\revise}[1]{{#1}}

\title{Align, Perturb and Decouple: Toward Better Leverage of Difference Information for RSI Change Detection}

\author{
Supeng Wang$^1$
\and
Yuxi Li$^{3}$\and
Ming Xie$^1$\and
Mingmin Chi$^{1,2,4}$\footnotemark[1]\and
Yabiao Wang$^{3}$\footnotemark[1]\and
Chengjie Wang$^{3,5}$\footnotemark[1]\and
Wenbing Zhu$^6$
\affiliations
$^1$School of Computer Science, Shanghai Key Laboratory of Data Science, Fudan University\\
$^2$Zhengzhou Zhongke Institute of Integrated circuit and System Application, China\\
$^3$YouTu Lab, Tencent, Shanghai\\
$^4$Zhongshan PoolNet Technology Ltd, China\\
$^5$Shanghai Jiao Tong University, China\\
 $^6$Rongcheer, China
\emails
spwang21@m.fudan.edu.cn,
\{mxie20,mmchi\}@fudan.edu.cn,
\{yukiyxli,caseywang,jasoncjwang\}@tencent.com,
louis.zhu@rongcheer.com
}

\begin{document}

\maketitle
\footnotetext[1]{Corresponding author}
\begin{abstract}
    Change detection is a widely adopted technique in remote sense imagery (RSI) analysis in the discovery of long-term geomorphic evolution. To highlight the areas of semantic changes, previous effort mostly pays attention to learning representative feature descriptors of \revise{a} single image, while the difference information is either modeled with simple difference operations or implicitly embedded via feature interactions. Nevertheless, such difference modeling can be noisy since it suffers from non-semantic changes and lacks explicit guidance from image content or context. In this paper, we revisit the importance of feature difference for change detection in RSI, and propose \revise{a} series of operations to fully exploit the difference information: Alignment, Perturbation and Decoupling (APD). Firstly, alignment leverages 
    contextual similarity to compensate for \revise{the} non-semantic difference in feature space. Next, a difference module trained with semantic-wise perturbation is adopted to learn more generalized change estimators, which reversely bootstraps feature extraction and prediction. Finally, a decoupled dual-decoder structure is designed to predict semantic changes in both content-aware and content-agnostic manners. Extensive experiments are conducted on benchmarks of LEVIR-CD, WHU-CD and DSIFN-CD, demonstrating our proposed operations \revise{bring} significant improvement and \revise{achieve} competitive results under similar comparative conditions. Code is available at https://github.com/wangsp1999/CD-Research/tree/main/openAPD
    
\end{abstract}

\section{Introduction}

\begin{figure}
    \centering
    \includegraphics[width=0.45\textwidth]{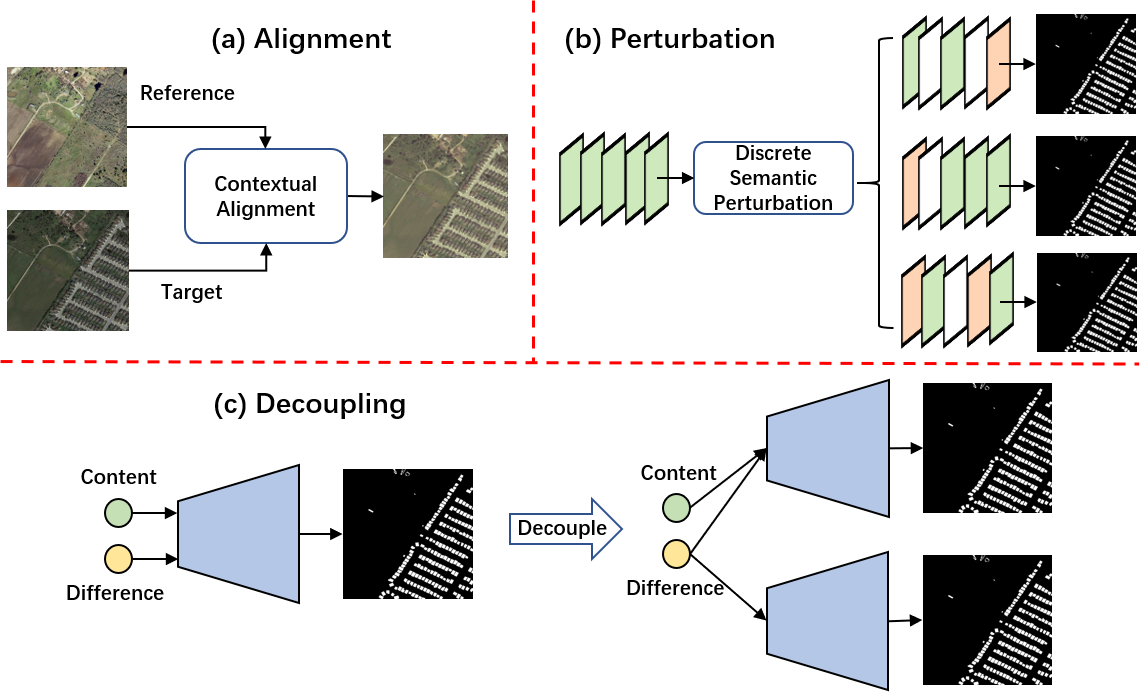}
    \caption{Conceptual illustration of operations proposed in this work to improve utilization of difference information. (a) A context reliant alignment operation to mitigate changes irrelevant to semantic. (b) A difference guidance module trained with discrete semantic perturbation. (c) Traditional single decoder with hybrid input is decoupled into a content-aware decoder and content agnostic decoder.}
    \label{fig:teaser}
\end{figure}

Change detection is a vision task aiming at identifying pixel-wise semantic changes between paired bitemporal images, this technique is helpful for analysis of remote sense imagery (RSI) with high resolution, which provides important information about changes in land surface and expansion of settlement during long time of observation~\cite{hussain2013change,daudt2018urban}.

With the development of remote satellite sensoring, there exist more open source 
\revise{databases} of remote sense imagery with fine-grained semantic annotations~\cite{levir,whu,dsifn}, which makes it possible to exploit the 
\revise{data-hungry} deep-learning approaches~\cite{alexnet,resnet} to achieve more accurate change detection. Due to the nature of pair-wise input and dense prediction, encoder-decoder architectures with siamese backbones prevail in recent \revise{efforts}, where features of single images are extracted separately and the difference information is exploited in a decoder to highlight areas of changing objects~\cite{FC,changeformer,BIT,changestar,SaDL}.

In a nutshell, recent approaches usually model the difference information between input pairs in simple and direct manners (e.g. taking difference or concatenation to obtain difference information)~\cite{FC,changeformer,changestar} or implicitly embed the difference into feature interaction~\cite{BIT,SaDL}, while still leaving some key issues open. \textbf{Firstly}, difference information is inherently vulnerable to pseudo-changes (e.g. the seasonal illumination changes during long-term observation), but few of previous works explicitly take such interference into account, hence these methods are not guaranteed to be robust enough to non-semantic changes. \textbf{Secondly}, most of previous literature take difference information from features for final decoding, ignoring the fact that difference naturally contains information of changeful objects, which reversely provides spatial guidance to representation learning~\cite{hussain2013change}. \textbf{Finally}, the relationship between image content and difference information is seldom discussed in prior works, the image content can be regarded as auxiliary prior information for change decoding, while also introducing some 
\revise{irrelevant change} cues distracting prediction.

With the reviews above, we claim that the difference information in current 
\revise{research} is still underutilized. Therefore, in this paper, we design series of operations aiming at mitigating the aforementioned issues and fully leveraging feature difference to boots change detection results. Concretely, we equip the hierarchical encoder-decoder network with three operations in sequential order: \textbf{A}lignment, \textbf{P}erturbation and \textbf{D}ecoupling (APD), which is illustrated in Figure~\ref{fig:teaser}. \textbf{Alignment:} To alleviate noise from pseudo-changes, we first propose a graph-based alignment module, which exploits the contextual similarity between patches to aggregate information from areas of the same semantic as compensation, this results in more precise extraction of semantic difference in following stages. \textbf{Perturbation:} Following the alignment operation, we propose a perturbation-aided difference prediction module. Especially, this module is trained with discrete semantic perturbation as feature augmentation, thus can recognize more generalized change patterns as guidance for feature extraction in \revise{the} following stages. \textbf{Decoupling:} We decouple the decoder into an asymmetric dual-stream structure for final prediction, one 
\revise{focuses} on integrating bitemporal contents with difference while the other takes pure difference information for decoding, this helps utilize the complementary property between image content and difference information while avoiding irrelevant noise.

We conduct experiments on three challenging benchmarks for RSI change detection: LEVIR-CD~\cite{levir}, WHU-CD~\cite{whu} and DSIFN-CD~\cite{dsifn}, and demonstrate the superiority of our proposed approach over existing competitive methods. 
\revise{Plenty} of ablation studies also verify the effectiveness of our proposed 
\revise{operations}. In a nutshell, the contribution of this paper can be summarized as following

\begin{itemize}
    \item We reconsider the problem of pseudo-changes and propose 
    \revise{a} graph-based alignment module to explicitly utilize contextual similarity for compensation.
    \item We propose a hierarchical difference extraction structure to guide the feature extraction process stage-by-stage, which is equipped with a specially designed discrete semantic perturbation scheme for feature augmentation.
    \item Different from most of prior works, we propose a dual decoder structure to decouple the utilization of image content from pure difference encoders.
    \item We integrate the proposed operation above and convert traditional encoder-decoder structure into a new change detector APD, which achieves competitive results on mainstream benchmarks.
\end{itemize}

\section{Related Works}
\subsection{Change Detection with Deep Learning}

We roughly divide deep learning-based change detection methods into two types~\cite{CDWork}: 
\revise{two-stage and one-stage methods}. In general, the two-stage method trains a CNN/FCN to classify the bitemporal images respectively, and compares their classification results to obtain change areas. To achieve this goal, both the bitemporal semantic labels and the change label should be provided~\cite{Two-stage1,Two-stage2,Two-stage3}.

The one-stage method \lyx{is a more prevailed framework in recent \revise{research}, which} takes change detection as a dense classification task \lyx{and} directly produces the change result from the bitemporal images. 
 \lyx{The FC series~\cite{FC} was one of the earliest methods to adopt convolution neural networks for change detection, where} three \lyx{architectures were proposed}: FC-EF, FC-Siam-Conc, and FC-Siam-Diff. FC-EF adopted the early fusion strategy, while FC-Siam-Conc and FC-Siam-Diff adopted the medium fusion strategy with different \lyx{difference} policies. 
\lyx{Besides, DTCDSCN~\cite{DTCDSCN} takes inspiration from the two-stage methods, and \revise{converts} change detection to multi-task pipeline with semantic map as auxiliary supervision. ChangeSTAR~\cite{changestar} further relaxes the paired requirement by taking unpaired images as input.}
\begin{figure*}[t]
    \centering
    \includegraphics[scale=0.49]{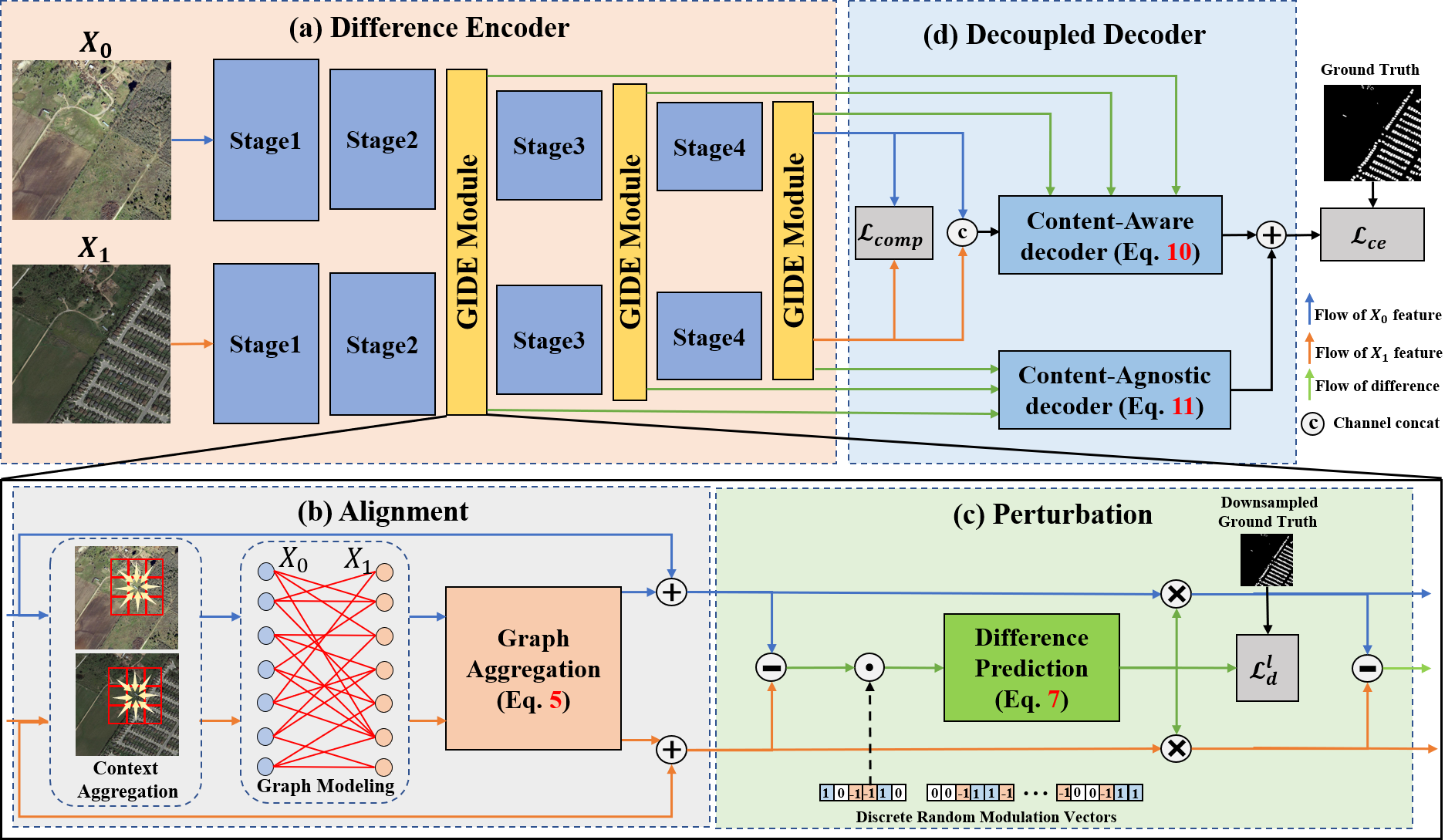}
    \caption{The overall framework of our proposed APD change detector. (a) The part of \revise{the} encoder equipped with \revise{our} proposed GIDE Module. (b) \revise{Detailed} illustration of graph-based alignment within GIDE. (c) Illustration of the perturbation guided difference prediction in GIDE. (d) The proposed \revise{asymmtetrical} duel decoder part for final prediction.}
    \label{fig:framework}
\end{figure*}

\subsection{Attention Mechanism in Change Detection}
\lyx{Although deep neural network achieves significant improvement over handcrafted features, traditional network structures can not explicitly capture the contextual reliance within RSI, thus can not accurately detect changes in object-level. Therefore, there appears works resorting to attention mechanism~\cite{SA,CA-SENet,vaswani2017attention} to help gather more contextual information.} 
Zhang et al.\cite{IFNet} proposed a deeply supervised change detection network (IFN) , which applies channel attention and spatial attention for feature enhancement. 

\lyx{Besides the contextual modeling, recent works also make attempt to integrate the attention mechanism into bitemporal interaction as an implicit replacement of feature difference operation}. Both BIT~\cite{BIT} and TransUNetCD~\cite{TransUNetCD} proposed a bitemporal image transformer by mixing CNN and transformer, \lyx{where the cross image interaction is conducted in latent encoded space}. ChangeFormer \cite{changeformer} \lyx{further developed} a siamese variant of SegFormer~\cite{xie2021segformer} for feature extraction. 


\subsection{Feature Augmentation}

\lyx{Feature augmentation aims at perturbing data in the encoded feature space to help \revise{the} neural network learn invariant representation. Different from data augmentation with low-level transformation, feature-level augmentation is regarded as increasing the semantic variance to ensure the generalization ability of models. ISDA~\cite{isda} perturbed the deep features in terms of their statistical characteristic, thus \revise{improving} the accuracy of recognition. The work of SFA~\cite{li2021simple} utilized a similar adaptive noise for domain generalization. Further, Manifold Mixup~\cite{manifoldmix} extended the widely-used mixup trick into feature space to smooth the semantic boundaries. MGD~\cite{yang2022masked} introduced random mask for knowledge distillation among deep features to obtain better results of student models.}

\lyx{In change detection, the difference information is usually derived from the discrepancy between two feature maps, which should be invariant to the difference order and channel-wise masking. Therefore, we take 
inspiration from feature augmentation methods and design a specialized discrete semantic perturbation mechanism to help difference prediction module recognize more general \revise{patterns}.}

\section{Approach}

\lyx{Figure~\ref{fig:framework} depicts the pipeline of our APD change detector. Similar to previous work, our method is built based on an encoder-decoder architecture. We denote} the bitemporal image pair as $\mc{X}_0$ and $\mc{X}_1$ $ \in \mathbb{R}^{3\times H\times W}$ with change label map $\mc{Y} \in \{0,1\}^{H\times W}$, \lyx{then} feed images \lyx{to a classic hierarchical deep neural network to extract feature representations of multi-level $\{\mc{F}^l_0, \mc{F}^l_1\}_l$}, where $l \in [2, N]$ represents the stage index (at most $N$ stages). \lyx{However, to fully leverage the difference information, we propose the concept of ``Align First and Then Difference'', and insert a \textbf{G}raph \textbf{I}nteractive and \textbf{D}ifference \textbf{E}nhancement module (GIDE) between stages in the encoder part, which} transforms $\mc{F}_{i}^{l}$ into more change-sensitive manifold $\widetilde{\mc{F}}_{i}^{l}$
\begin{equation}
\begin{aligned}
    \widetilde{\mc{F}}_{0}^{l}, \widetilde{\mc{F}}_{1}^{l}, \mc{O}^{l} & = \textbf{GIDE}(\mc{F}_{0}^{l}, \mc{F}_{1}^{l}) = P\left(A\left(\mc{F}_{0}^{l}, \mc{F}_{1}^{l}\right)\right) \\
    \mc{F}_{0}^{l+1}, \mc{F}_{1}^{l+1} & = \textbf{Enc}_l\left( \widetilde{\mc{F}}_{0}^{l}, \widetilde{\mc{F}}_{1}^{l} \right),
\end{aligned}
\end{equation}
where $ \textbf{Enc}_l\left( \cdot, \cdot \right)$ denotes the $l$-th stage of \revise{the encoder}. Specifically, GIDE is composed of an alignment module $\mathit{A} \left (\cdot, \cdot \right ) $, which aggregates pixel-level semantic information, and a perturbation-aided difference module $\mathit{P} \left (\cdot, \cdot \right ) $ \revise{which} augments feature difference and highlights local areas. Additional to the alignment and perturbation, $\mathit{P} \left (\cdot, \cdot \right ) $ also generates feature difference $\mc{O}^l$ as pure difference information for \revise{the following} decoding. \lyx{In the part of \revise{the decoder}}, we decouple the \lyx{classical single} decoder into a complementary structure of content-aware decoder $\mathit{D}_{aw}$ and content-agnostic decoder $\mathit{D}_{ag}$ to excavate the content and difference information respectively \lyx{for final prediction}.

\subsection{Context-aware Alignment}
\lyx{The alignment operation aims at alleviating \revise{the disturbance} of pseudo-changes and amplifying \revise{the area} of semantic change, thus we resort to an undirected bipartite graph as \revise{an aggregation} structure to align features (as shown in Figure~\ref{fig:framework} (b)).}

\lyx{Formally, suppose both $\mc{F}_{0}^{l}, \mc{F}_{1}^{l}$ are of resolution $H_l \times W_l$, then we can take each pixel in feature map as a graph node and build an adjacent matrix $\mc{G}$, which is a bipartite graph, i.e. an edge can only link pixels from different feature maps, thus the adjacent matrix can be represented as four-dimensional, $ \mc{G} \in \{0, 1\}^{H_l \times W_l\times H_l \times W_l}$}, and the node feature is the feature of \revise{the} corresponding pixel. Considering the feature of \revise{a} single pixel can be sensitive \revise{to} non-semantic changes, while its contextual information can be more robust since it contains more structural information, \revise{therefore} it is more suitable to encode the edge between nodes with the reference of contextual similarity. To do this, we first obtain the context feature $\mc{F}_{0,c}^{l},\mc{F}_{1,c}^{l}$ via a non-parameterized dilated convolution of dilate factor $d$ and kernel size $k$, for $m=0,1$, we have
\begin{equation}
\small
    \mc{F}_{m,c}^{l}[u,v] = \sum_{(i,j)\in [-k, k]^2 \atop (i,j)\neq (0,0)}{\mc{F}_{m}^l[u + i d, v + j d],}
\end{equation}
\lyx{where $[u,v]$ indicates obtaining the pixel-wise feature at location $(u, v)$. To build graph $\mc{G}$, for each pixel $(u,v)$ on $ \mc{F}_{0, c}^{l}$, we compute top $n$ nearest coordinates on $ \mc{F}_{1, c}^{l}$ }

\begin{equation}
\small
    \mc{H}_{u,v} = \textbf{TopK}_{i\in [0, H_l], j\in [0, W_l]}\left({\left \| \mc{F}_{0,c}^{l}[u,v] - \mc{F}_{1,c}^{l}[i, j] \right \| }_{2}\right),
\end{equation}
with the grouped coordinate set of nearest neighbor $\mc{H}_{u,v}$, the high-dimensional adjacent matrix can be expressed as
\begin{equation}
    \mc{G}[u,v,i,j]=\left\{\begin{array}{cc}
1, & \text { if } (i, j) \in \mc{H}_{u,v}  \\
0, & \text { otherwise. }
\end{array}\right.
\end{equation}
\lyx{With the relation graph $\mc{G}$ based on context, we can take the naive graph convolution to aggregate information from similar node pairs to both feature maps \revise{as follows}}
\begin{equation}
\begin{aligned}
    {\mc{F}_{0}^l}', {\mc{F}_{1}^l}' & = \textbf{GCN}(\mc{G}, \mc{F}_{0}^l, \mc{F}_{1}^l) \\
    \hat{\mc{F}}_{0}^l = \mc{F}_{0}^l & + {\mc{F}_{0}^l}' \quad  \hat{\mc{F}}_{1}^l = \mc{F}_{1}^l + {\mc{F}_{1}^l}',
\end{aligned}
\label{eq:graph_conv}
\end{equation}
\lyx{where $\hat{\mc{F}}_{0}^l, \hat{\mc{F}}_{1}^l$ are aligned features after aggregation of graph convolution and will be utilized in following difference prediction module.}


\subsection{Perturbation-aided Difference Prediction}
\label{sub:perturbation}
\lyx{After the feature alignment, we make \revise{attempts} to leverage \revise{the aligned feature} to produce coarse spatial guidance, thus \revise{helping} feature extraction in \revise{the following stage}. To this end, we introduce a feature augmentation mechanism to train a coarse difference prediction module (as \revise{described} in Figure~\ref{fig:framework} (c)).}

\lyx{We start with a simple difference operation, however, we explicitly modulate the channel of feature difference with a random vector $\mathbf{v}$ filled with discrete values}
\begin{equation}
\begin{aligned}
    \hat{\mc{O}}^l & = (\hat{\mc{F}}_{0}^l - \hat{\mc{F}}_{1}^l) \odot \mathbf{v} \\
    \textbf{s.t.} \quad \|\mb{v}\|_1 & = (1-\tau) C^l \quad \mathbf{v} \in \{+1, -1, 0\}^{C^l},
\end{aligned}
\end{equation}
\lyx{where $C^l$ is the channel dimension of $\hat{\mc{F}}_{0}^l, \hat{\mc{F}}_{1}^l$, and $\odot$ represents channel-wise modulation, i.e. the $i$-th element of vector $\mathbf{v}$ is broadcasted and multiplied to all pixels on the $i$-th channel. The value of vector $\mathbf{v}$ is constrained within $0$ and $\pm 1$, thus each channel of feature difference is either reversed ($-1$), masked ($0$) or retained ($+1$). The insight behind such perturbation is that even though part of semantic information is masked out or reversed, the preserved semantic should be informative enough to coarsely highlight the change. Besides, to prevent too many channels from being masked, we define a mask ratio $\tau \in [0,1]$ to control the ratio of $0$ values in $\mathbf{v}$. It should \revise{be noted} that \revise{channel-wise} modulation is only adopted during training and removed from inference.}

\lyx{The modulated difference $\hat{\mc{O}}^l$ is fed into an ASPP~\cite{chen2018encoder}-like structure to obtain a \revise{one-dimensional} coarse mask $\mc{M}^l \in \mathbb{R}^{H_l \times W_l}$}
\begin{equation}
\small
    \mc{M}^l = \sigma\left(\textbf{MLP}\left(\textbf{Cat}\left(\textbf{GAP}(\hat{\mc{O}}^l), \hat{\mc{O}}^l \right)\right)\right),
\end{equation}
\lyx{where $\textbf{GAP}(\cdot)$ denotes global average pooling and $\textbf{Cat}(\cdot, \cdot)$ indicates concatenation between two \revise{features} along channel dimension, note that the pooled feature is broadcasted to all pixels to align the resolution, $\textbf{MLP}(\cdot)$ represents multi-layer perceptron mapping the input to single channel mask, and a sigmoid function $\sigma(\cdot)$ is adopted to obtain coarse-level change area in $l$-th stage. In the end, GIDE} feeds back such difference-dominant information to the original \revise{siamese-aligned} feature
\begin{equation}
    \widetilde{\mc{F}}^l_{0} = \hat{\mc{F}}^l_{0} \otimes \mc{M}^l \quad \widetilde{\mc{F}}^l_{1} = \hat{\mc{F}}^l_{1} \otimes \mc{M}^l \quad \mc{O}^l = (\hat{\mc{F}}_{0}^l - \hat{\mc{F}}_{1}^l) \otimes \mc{M}^l,
\end{equation}
\lyx{where $\otimes$ indicates \revise{spatial-wise} modulation, i.e. $\mc{M}^l$ is broadcast to different channels.}

\noindent\textbf{Deep Supervision.} Besides, we inject additional supervised \revise{objectives} in the perturbation module as deep supervision to \lyx{ensure the accuracy of estimated guidance mask}. \lyx{Concretely,} we downsample \revise{the original} change map \lyx{$\mc{Y}$} to adapt \lyx{the size of mask $\mc{M}^l$ as $\mc{Y}^l$} and apply the binary cross-entropy as loss
\begin{equation}
\begin{aligned}
    \mc{L}_{d}^l = & - \frac{1}{H_lW_l}\sum_{(i,j)}{\mc{Y}^l[i,j] log (\mc{M}^l[i,j])} \\ 
                   & - \frac{1}{H_lW_l}\sum_{(i,j)}{(1 - \mc{Y}^l[i,j]) log (1 - \mc{M}^l[i,j])}.
\end{aligned}
\label{eq:deepsupervision}
\end{equation}
\lyx{The deep supervision together with our designed random perturbation helps train more generalized \revise{modules} to differentiate various change patterns.}

\subsection{Decouple Decoders}
\lyx{In the decoder part, to exploit the complementary property of image content and pure difference, we devise an asymmetric dual-decoder structure, which takes $ {\mc{F}}^N_{0}, {\mc{F}}^N_{1}$ and $\mc{O}^l$ as input and \revise{predicts} the change areas.}


The content-aware decoder \lyx{takes the encoded image features as input, and hierarchically \revise{appends} difference information to generate the intermediate feature, finally a segmentation head is applied on \revise{the final} output to obtain \revise{the prediction}}
\begin{equation}
\begin{aligned}
    \mc{D}_{aw}^N & = \textbf{MLP}\left(\textbf{Cat}({\mc{F}}^N_{0},{\mc{F}}^N_{1})\right) \\
    \mc{D}_{aw}^{l-1} & = \textbf{Dec}_{l, aw}\left(\mc{S}_{\uparrow}(\mc{D}_{aw}^l), \mc{O}^{l-1}\right) \\
    \hat{\mc{Y}}_{aw} & = \mc{T}_{aw}(\mc{D}_{aw}^1),
\end{aligned}
\end{equation}
where $\textbf{Dec}_{l, aw}(\cdot, \cdot)$ is the $l$-th decoding block of decoder, which consists of concatenation and two cascaded Conv-BN-ReLU blocks with kernel size $3\times 3$, $\mc{S}_{\uparrow}(\cdot)$ represents the upsampling operation, and $\mc{T}_{aw}(\cdot)$ is the segmentation head. \lyx{As for the content-agnostic decoder, we only feed the pure difference information as input, the other structure is similar to the content-aware one}
\begin{equation}
\begin{aligned}
    \mc{D}_{ag}^N & = \textbf{MLP}\left(\mc{O}^N\right) \\
    \mc{D}_{ag}^{l-1} & = \textbf{Dec}_{l, ag}\left(\mc{S}_{\uparrow}(\mc{D}_{ag}^l), \mc{O}^{l-1}\right) \\
    \hat{\mc{Y}}_{ag} & = \mc{T}_{ag}(\mc{D}_{ag}^1),
\end{aligned}
\end{equation}
where the structure of $\textbf{Dec}_{l, ag}(\cdot, \cdot)$ is similar to that of content-aware decoder but the concatenation is replaced by summation of two input features. Finally, both output are summed up with an activation function to obtain final results
\begin{equation}
    \hat{\mc{Y}} = \sigma(\hat{\mc{Y}}_{ag} + \hat{\mc{Y}}_{aw}).
\end{equation}



\subsection{Loss Function}
\lyx{During training, we simply take the cross-entropy $\mc{L}_{ce}$ to supervise the final output $\hat{\mc{Y}}$ and the total loss function can be expressed as Equation~(\ref{eq:loss}) with balance factor $\lambda_1$ and $\lambda_2$}
\begin{equation}\label{eq:loss}
    \mc{L}_{total} = \mc{L}_{ce} + \lambda_{1}\sum_{l}\mc{L}^l_{d} + \lambda_{2}\mc{L}_{comp}.
\end{equation}
\lyx{In Equation~(\ref{eq:loss}), we introduce an additional comparative loss $\mc{L}_{comp}$ for feature regularization, which is expressed as}
\begin{equation}
\small
    \begin{aligned}
    \mc{L}_{comp} = & \frac{1}{HW}\sum_{(i,j)}{\mc{Y}^N[i,j] \left[\left\| \widetilde{\mc{F}}^N_{0}[i,j] - \widetilde{\mc{F}}^N_{1}[i, j] \right\|_2 - \gamma\right]_+} \\
                 + & \frac{1}{HW}\sum_{(i,j)}{(1 - \mc{Y}^N[i,j]) \left\| \widetilde{\mc{F}}^N_{0}[i,j] - \widetilde{\mc{F}}^N_{1}[i, j] \right\|_2},
\end{aligned}
\end{equation}
\lyx{where $\left[\cdot\right]_+$ represents clipped by $0$ if the value inside is negative and $\gamma$ is a hyperparameter. This comparative term helps backbone to distinguish feature of changed objects.}


\begin{table*}[t]
\centering
\scalebox{0.76}{
\begin{tabular}{c|c|cccc|cccc|cccc}
\toprule
\multirow{2}{*}{Method} &\multirow{2}{*}{Backbone} & \multicolumn{4}{c}{LEVIR-CD} &\multicolumn{4}{c}{DSIFN-CD}&\multicolumn{4}{c}{WHU-CD}\\
&& Precision & Recall & F1 & IoU & Precision & Recall & F1 & IoU & Precision & Recall & F1 & IoU\\ 
\midrule
FC-EF & UNet & 86.91 & 80.17 & 83.40 & 71.53 & 72.61 & 52.73 & 61.09 & 43.98 & 71.63 & 67.25 & 69.37 & 53.11\\
FC-Siam-Diff & UNet & 89.53 & 83.31 & 86.31 & 75.92 & 59.67 & 65.71 & 62.54 & 45.50 & 47.33 & 77.66 & 58.81 & 41.66\\
FC-Siam-Conc & UNet & 91.99 & 76.77 & 83.69 & 71.96 & 66.45 & 54.21 & 59.71 & 42.56 & 60.88 & 73.58 & 66.63 & 49.95\\
SNUNet & UNet++ & 89.18 & 87.17 & 88.16 & 78.83 & 60.60 & 72.89 & 66.18 & 49.45 & 85.60 & 81.49 & 83.50 & 71.67\\
DTCDSCN & SE-Res34 & 88.53 & 86.83 & 87.67 & 78.05 & 53.87 & 77.99 & 63.72 & 46.76 & 63.92 & 82.30 & 71.95 & 56.19\\
STANet & ResNet18 & 83.81 & \textbf{91.00} & 87.26 & 77.40 & 67.71 & 61.68 & 64.56 & 47.66 & 79.37 & 85.50 & 82.32 & 69.95\\
BiT & ResNet18 & 89.24 & 89.37 & 89.31 & 80.68 & 68.36 & 70.18 & 69.26 & 52.97 & 86.64 & 81.48 & 83.98 & 72.39\\
TransUNetCD & ResNet50 & 92.43 & 89.82 & 91.11 & 83.67 & 71.55 & 69.42 & 66.62 & 57.95 & 93.59 & 89.60 & 93.59 & 84.42\\
ChangeFormer & MiT-B1 & 92.05 & 88.80 & 90.40 & 82.48 & 88.48 & 84.94 & 86.67 & 76.48 & 89.12 & 82.73 & 85.61 & 75.14\\

\midrule
Ours & ResNet18 & \textbf{92.81} & 90.64 & \textbf{91.71} & \textbf{84.69} & \textbf{89.39} & \textbf{86.40} & \textbf{87.87} & \textbf{78.36} & \textbf{95.10} & \textbf{95.26} & \textbf{95.18} & \textbf{90.80} \\
\bottomrule
\end{tabular}}
\caption{Quantitative \revise{comparison} results of different change detection methods on LEVIR-CD, DSIFN-CD and WHU-CD}
\label{tab:MainResult}
\end{table*}

\begin{figure*}
    \centering
    \includegraphics[scale=0.51]{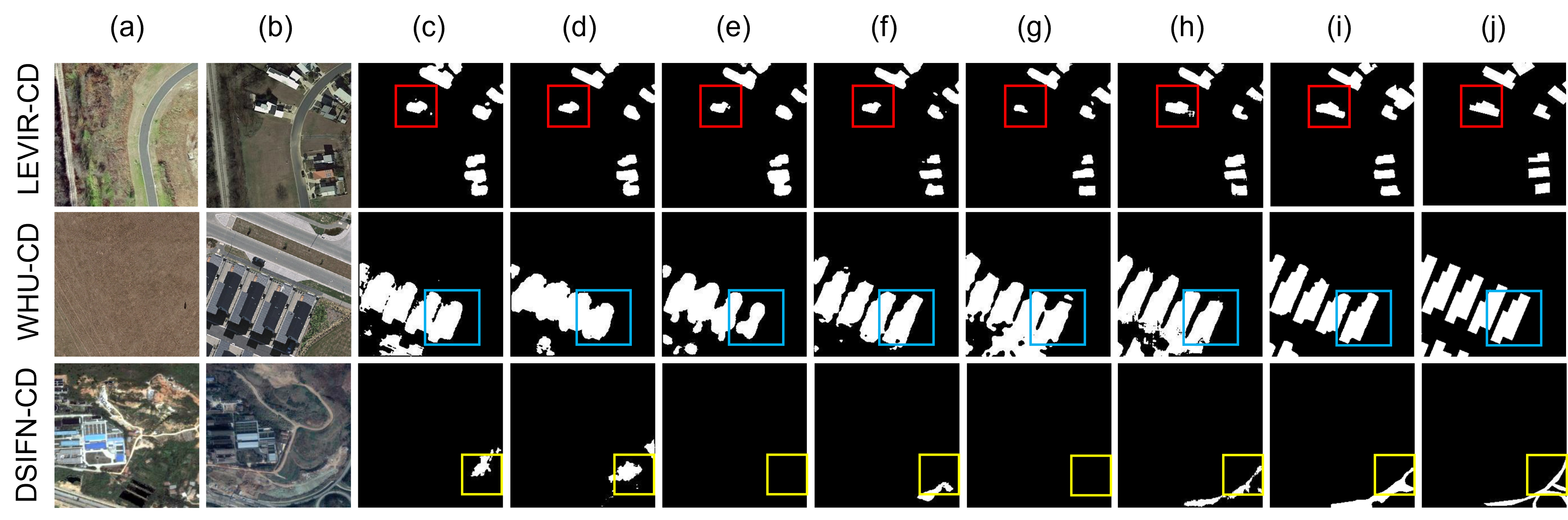}
    \caption{Qualitative results of different CD methods on LEVIR-CD, DSIFN-CD and WHU-CD dataset: (a) Pre-change image, (b) Post-change image, (c) FC-EF, (d) FC-Siam-Diff, (e) FC-Siam-Conc, (f) DTCDSCN, (g) BIT, (h) ChangeFormer, (i) Ours, and (j) Ground-truth.}
    \label{fig:quaResult}
\end{figure*}

\section{Experiment}
\subsection{Experiment Setup}
\noindent\textbf{Dataset.} 
We \lyx{evaluate} our proposed \lyx{APD change detector} on three \lyx{publicly} change detection datasets: LEVIR-CD~\cite{levir}, WHU-CD~\cite{whu} and DSIFN-CD~\cite{dsifn}. LEVIR-CD is a public large building \lyx{change detection} dataset that contains 637 bitemporal RS image pairs of resolution 1024×1024. We utilize the default train/val/test \lyx{split}. WHU-CD is \lyx{another} public building \lyx{change detection} dataset, \lyx{which} consists of one pair of ultra-resolution (0.075m) aerial \revise{images} of size 32507×15354. We follow the default \lyx{cropping policy} of size 512×512 and dataset split(train/test) provided by the authors. DSIFN-CD contains the changes in six major cities’ landcover objects in China. We divide the 512×512 images into 256×256 pixel patches without overlapping, and we follow the default standard train/val/test split. \lyx{Consequently}, there are 14400/1360/192 samples for training/val/test.

\noindent\textbf{Implementation Details.}
We implemented our model under the Pytorch framework, using a single NVIDIA GeForce GTX 1080 Ti GPU for training and the batch size is set to 8. During training, we 
apply data augmentation through random flip, crop and photometric distortion. We use AdamW with weight decay equal to 0.05 \lyx{for optimization}. The initial learning rate is 0.001 and we train models for 60k, 40k and 100k iterations for LEVIR-CD, WHU-CD and DSIFN-CD \revise{datasets}. For context aggregation, we set $k=1$ and $d=16$. The hyperparameters in loss term are $\lambda_1=1.0, \lambda_2=1.0, \gamma=1.0$.

\noindent\textbf{Evaluation Metrics.}
\lyx{For horizontal comparison with other methods, we follow the common setting and} use the F1 \lyx{score} and Intersection over Union (IoU) with regard to the change \lyx{objects} as the primary evaluation indices. Meanwhile, we \lyx{also} report precision(P) and recall(R) of the change \lyx{objects}.

\subsection{Main Result}
\noindent\textbf{Methods for Comparison.}
To verify the effectiveness of our method, we make comparison with several \lyx{advanced change detection approaches}, including three purely convolutional-based methods (FC-EF~\cite{FC}, FC-Siam-Conc~\cite{FC}, FC-Siam-Diff~\cite{FC}), three attention-aided methods (DTCDSCN~\cite{DTCDSCN}, STANet~\cite{STANet}, SNUNet~\cite{SNUNet}) and three methods \lyx{with transformer-like structure} (BIT~\cite{BIT}, ChangeFormer~\cite{changeformer}, TransUNetCD~\cite{TransUNetCD}).


\noindent\textbf{Quantitative Results.} Table~\ref{tab:MainResult} reports the overall quantitative comparison results on LEVIR-CD, DSIFN-CD and WHU-CD. In the \lyx{datasets of} DSIFN-CD and WHU-CD, \lyx{our proposed APD change detector} outperforms other methods, reaching the best level in \lyx{terms of} all four metrics. In the LEVIR-CD dataset, although our method does not achieve the best precision, \lyx{APD can still detect} more changing pixels, which makes significant advantages in the other three metrics. For example, the F1 score of our method exceeds the latest ChangeFormer by 1.31\%, 1.2\%, and 9.57\% on the three datasets, respectively. In Table~\ref{tab:MainResult}, we also \lyx{indicate the information of \revise{the utilized} backbone} of different CD methods. It can be \lyx{observed} that \lyx{APD only applies a simple and lightweight ResNet-18 network as \revise{a} feature extractor} and \lyx{does} not use complex structures such as UNet\cite{FC,SNUNet} or transformer-based network\cite{changeformer}, which are widely used in segmentation tasks. \lyx{On the other hand, although our approach only takes ResNet18 as \revise{the backbone}, it can still \revise{outperform} competitors with larger model capacity (ResNet50) or more advanced structure (MiT-B1), which indirectly manifests the superiority of our proposed GIDE module and decoupled decoders}.

\noindent\textbf{Qualitative Results.} In addition, Figure~\ref{fig:quaResult} also shows the visualization comparison of the different change detection methods on the three datasets. As highlighted in red, blue and yellow respectively, \lyx{our} proposed method captures more detailed change \lyx{compared} with other \lyx{change detection schemes}. In the visualization results of LEVIR-CD dataset, our APD detector can not only detect the building change \lyx{more accurately}, but also avoid some noise (e.g. changes in the appearance of land cover, seasonal illumination changes, etc.) that affects the contour of the changed area. From the visualization results of the WHU-CD dataset, it can be seen that \lyx{most} compared methods cannot eliminate the \revise{pseudo-change} caused by shadows when detecting changes. \lyx{In contrast}, our method can eliminate such \revise{pseudo-change}, which proves that our method can learn effective context, eliminate the irrelevant change and express the real semantic variation. At the same time, our method can also effectively detect subtle changes, which \revise{are} \lyx{observed} from the visualization results of the DSIFN-CD dataset. The compared methods can hardly detect the detailed changed area of the long and narrow road due to the lack of precise semantic difference information. However, our APD detector can effectively capture the subtle variation and \lyx{generate} \revise{a more} precise change map, which verifies the \lyx{superiority} of \lyx{APD detector}.

\begin{table}[t]
\centering
\scalebox{0.73}{
\begin{tabular}{ccc|cccc}
\toprule
\multirow{2}{*}{Align} & \multirow{2}{*}{Perturb} & \multirow{2}{*}{Decouple} & \multicolumn{4}{c}{LEVIR-CD}\\ 
& & &  Precision & Recall & F1 & IoU \\ 
\midrule
   &   &   & 85.59 & 86.27 & 85.93 & 75.33\\
\checkmark  &   &   & 87.06 & 87.16 & 87.11 & 77.16\\
   & \checkmark  &   & 87.92 & 87.43 & 87.68 & 78.06\\
   &   & \checkmark  & \textbf{93.83} & 87.31 & 90.45 & 82.56\\
\checkmark  & \checkmark &  & 92.07 & 90.02 & 91.03 & 83.54\\
\checkmark  &   & \checkmark  & 93.00 & 89.50 & 91.22 & 83.36\\
   & \checkmark  & \checkmark  & 93.38 & 89.27 & 91.28 & 83.95\\
\checkmark  & \checkmark  & \checkmark  & 92.81 & \textbf{90.64} & \textbf{91.71} & \textbf{84.69}\\
\bottomrule
\end{tabular}}
\caption{Ablation study on the effectiveness of operations proposed in our approach.}
\label{tab:AblationStudy1}
\end{table}

\subsection{Ablation Study}
\label{sec:ablation}
\noindent\textbf{Verification Experiment on Proposed Operations}.
We design ablation experiments for each \lyx{proposed \revise{operation}, i.e. Alignment, Perturbation and Decouple. For each operation, we provide \revise{a} specific baseline counterpart to demonstrate the effectiveness}. 
\lyx{The overall results are} shown in Table~\ref{tab:AblationStudy1}. The experiment shows that the Alignment module, Perturbation-aided difference module and Decoupled Decoder are helpful for \lyx{change detection performance}.

\begin{figure}[t]
    \centering
    \includegraphics[scale=0.38]{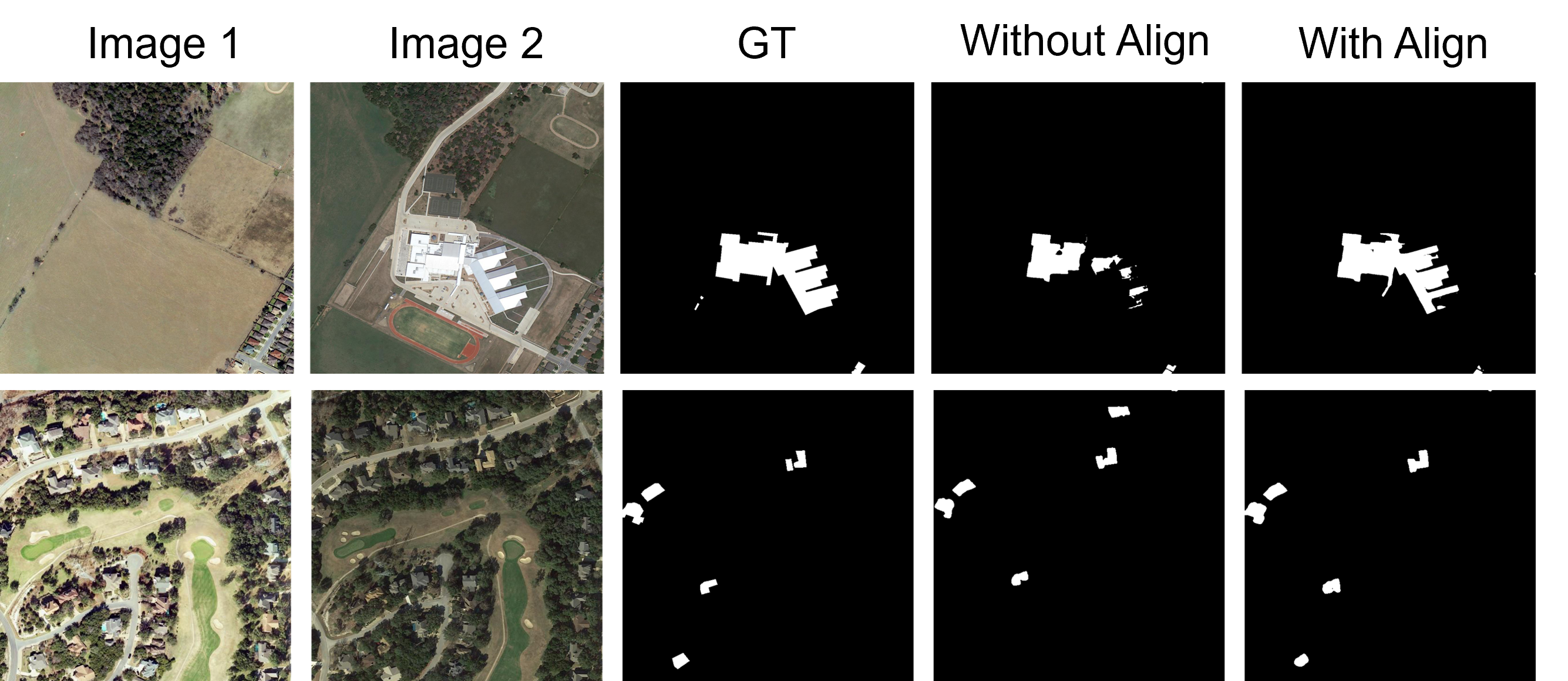}
    \caption{Visualization results of effect from Alignment}
    \label{fig:Alignment}
\end{figure}

\lyx{To evaluate the alignment operation, we remove the context-aided alignment in GIDE on purpose and \revise{directly} fed the siamese features into difference prediction. In terms of Table~\ref{tab:AblationStudy1}, the context-aided alignment significantly improves the recall and IoU score.} This is because the Alignment module \lyx{effectively utilizes the contextual similarity to gather} information from areas of the same semantic to enhance the semantic difference information. \lyx{This \revise{improvement} is also reflected} in Figure~\ref{fig:Alignment}, if the alignment part in the GIDE module is removed, the noise caused by the lack of the contextual information makes the model unable to distinguish between real changes and pseudo-changes, thus affecting the precise recognition of the changed areas.

\begin{figure}[t]
    \centering
    \includegraphics[scale=0.378]{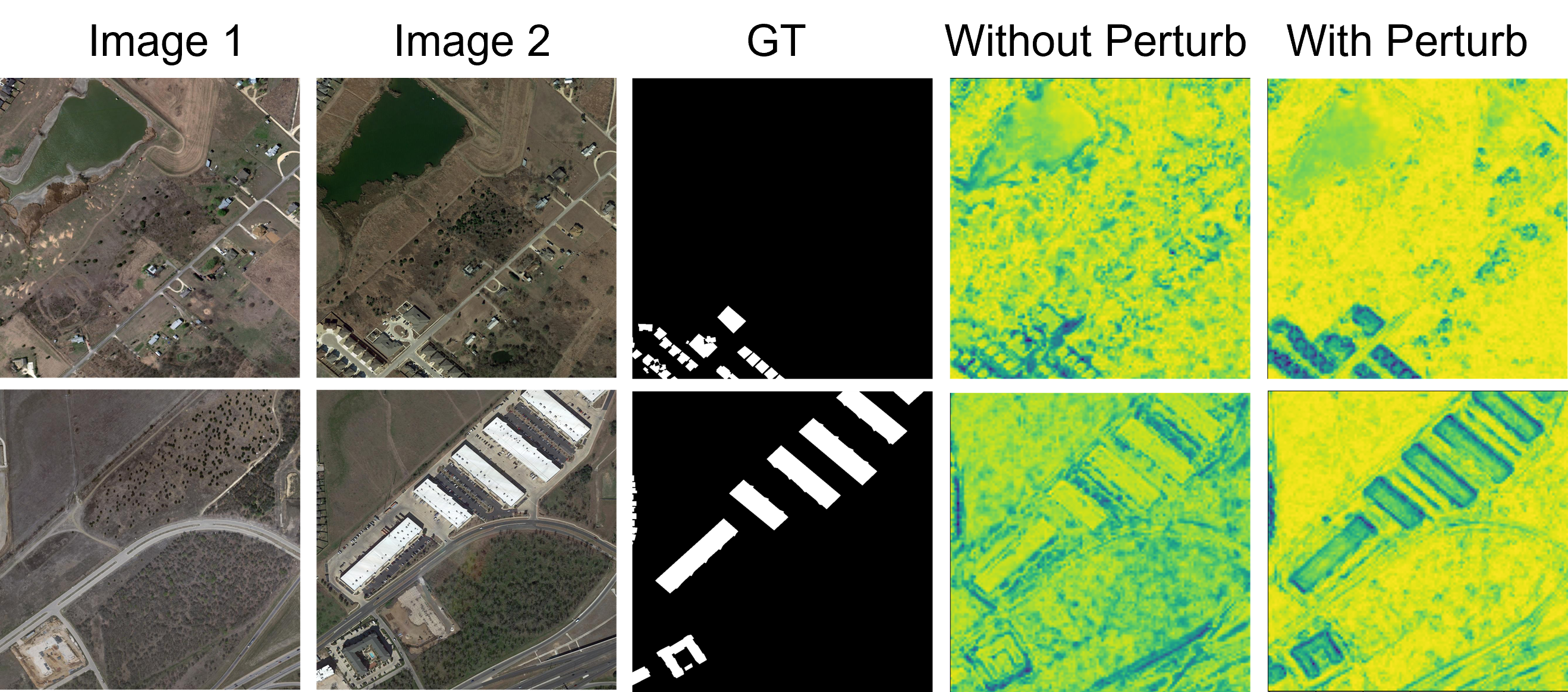}
    \caption{Feature visualization results of effect from Perturbation}
    \label{fig:Perturb}
\end{figure}

\lyx{To verify the effectiveness of perturbation, we also devise a counterpart by eliminating the deep supervision and semantic perturbation, instead we directly take the feature from both images and their difference as output. Table~\ref{tab:AblationStudy1} demonstrates that our proposed perturbation-aided deep supervision enssentially improve the recall and F1-score, since} combination between deep supervision and random disturbance can help difference predictor focus on more general patterns of changeful objects, consequently the estimated change area provides in-depth guidance for following feature extraction. \lyx{In additional}, we \lyx{also visualize the heatmap of intermediate feature to show the impact from perturbation}. Figure~\ref{fig:Perturb} shows the feature visualization before and after \lyx{introducing} the perturbation in the second stage of the backbone. It can be \lyx{observed} that with the perturbation-aided module, our model significantly reduces the impact of the pseudo-change (pond, road, etc.) and strengthens focus on the changeful objects.

\begin{figure}[t]
    \centering
    \includegraphics[scale=0.33]{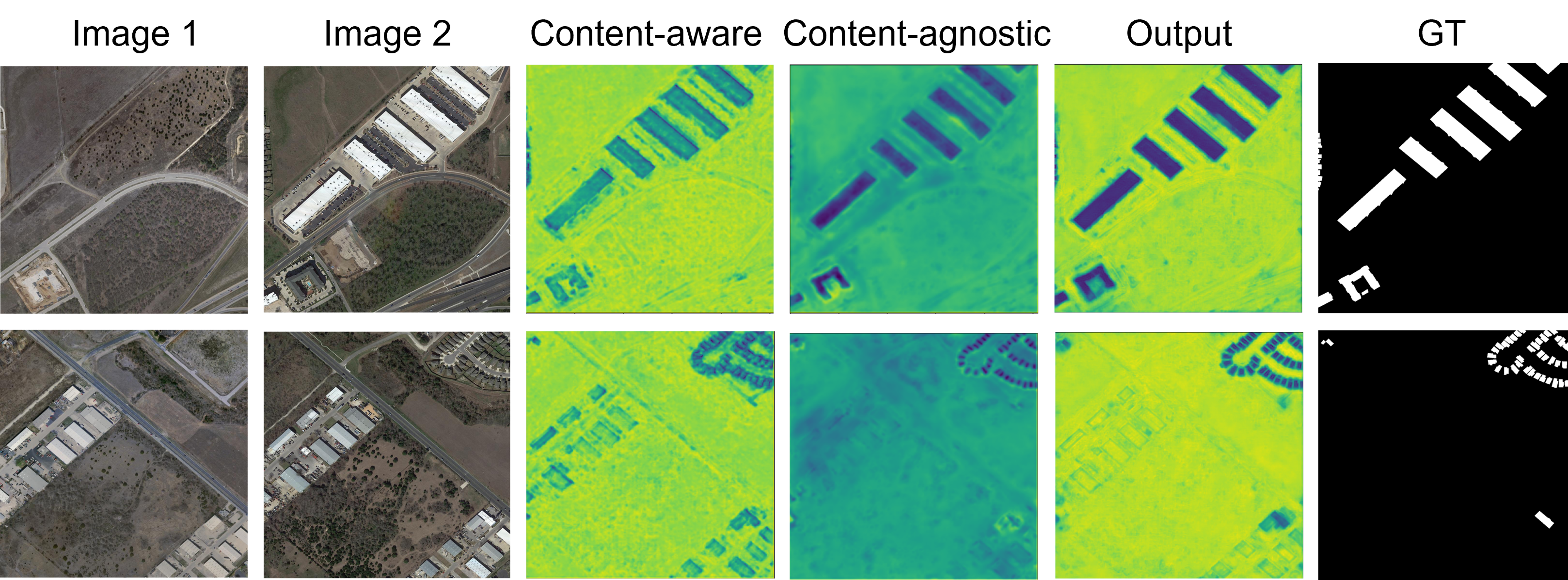}
    \caption{Feature visualization of effect from Decoupled Decoder}
    \label{fig:Decoupled}
\end{figure}

\lyx{Finally, we evaluate the impact from decoupled decoder, specifically, to build the baseline, we introduce a single decoder structure which takes the concatenation of image features and difference information as input. From the Table~\ref{tab:AblationStudy1}, we find decoupled structure brings substantial performance gain on detection precision and IoU score. This demonstrates that} decoupled decoders make full use of \lyx{the complementary between} image content and difference information for better change detection. Figure~\ref{fig:Decoupled} shows the feature visualization of two shunts in our decoupled decoders. 
\lyx{We observe} that the output of the Content-agnostic Decoder \lyx{can retains the profile of real semantic change}. \lyx{Further, the dual-decoder structure can help} detect \lyx{some small and detailed} changes that could not be detected by pure difference information.


\begin{table}[t]
\centering
\scalebox{0.73}{
\begin{tabular}{c|c|cccc}
\toprule
\multirow{2}{*}{Deep Supervision} &\multirow{2}{*}{Mask Ratio} & \multicolumn{4}{c}{LEVIR-CD}\\
&& Precision & Recall & F1 & IoU \\ 
\midrule
\multirow{4}{*}{DiceLoss}
& 1/2 & 86.78 & 86.89 & 86.78 & 76.74 \\
& 1/4 & 87.65 & 87.23 & 87.44 & 77.68 \\
& 1/8 &  86.64  & 87.37 & 87.00 & 76.99 \\
& 1/16 & 86.80 & 87.41 & 87.10 & 77.15 \\
\midrule
\multirow{4}{*}{CrossEntropyLoss}
& 1/2 & 87.51 & 87.5 & 87.5 & 77.78 \\
& 1/4 & \textbf{87.92} & 87.43 & \textbf{87.68} & \textbf{78.06} \\
& 1/8 &  87.60  & \textbf{87.5} & 87.55 & 77.86 \\
& 1/16 & 86.90 & 87.35 & 87.12 & 77.18 \\
\bottomrule
\end{tabular}}
\caption{Ablation study on deep supervision and perturbation ratios used in the perturbation operation}
\label{tab:AblationStudy2}
\end{table}

\noindent\textbf{Verification Experiment on Hyperparameters.}
\lyx{Next we also conduct parameter verification experiments to test sensitivity to hyperparameters} in the disturbance module on the LEVIR-CD dataset. \lyx{The results are} shown in Table~\ref{tab:AblationStudy2}.

First \lyx{we evaluate} the effect of the ratio of the \lyx{masked channels, conceretly}, we set the ratio $\tau$ to 50\%, 25\%, 12.5\% and 6.25\% of the input feature dimension. The experimental results show that \lyx{there exist negative impact on detection performance when the perturbation ratio is} too high or too low, this is because \lyx{large perturbation ratio can result in too many masked channels and loss in difference information, thus the difference prediction underfit to underlying change patterns and} affect feature extraction in following stages. \lyx{On the other hand,} too few \lyx{perturbed channels will decrease the variance in difference information, thus the difference prediction overfit to specific change patterns}. Therefore, we choose 25\% as the most appropriate ratio. 

Secondly, \lyx{we evaluate \revise{the effect} of different forms of} deep supervision $\mc{L}^l_d$. \lyx{From Table~\ref{tab:AblationStudy2}, when replacing cross-entropy with dice loss, the} results \lyx{\revise{are} degraded} in different ratios of \lyx{perturbation}, because small objects account for a certain proportion in the change detection dataset, \lyx{to which} the dice loss is very sensitive. \lyx{Therefore,} once some pixels of small objects are incorrectly predicted, it will lead to large \lyx{fluctuation} in the \lyx{backward gradient}. \lyx{Instead, the cross-entropy takes each pixel equally, which alleviates} this problem.

\section{Conclusion}

In this paper, we review the issue of underutilization of difference information in previous methods of RSI change detection, and propose a new change detector termed as APD. The APD change detector features with three elaborately designed operations, i.e. Alignment, Perturbation and Decoupling, to fully leverage the difference information and bootstrap change detection results. With a lightweight backbone, our APD detector can effectively improve the performance on challenging benchmarks of change detection and 
\revise{achieve} state-of-the-art results in most metrics, \revise{and} extensive ablation studies also verify the effectiveness of each operation.

\section*{Acknowledgements}
This work was supported in part by Natural Science Foundation of China under contract 62171139, and in part by Zhongshan science and technology development project under contract 2020AG016.

\bibliographystyle{named}
\bibliography{ijcai23}

\begin{thebibliography}{}

\bibitem[\protect\citeauthoryear{Bandara and Patel}{2022}]{changeformer}
Wele Gedara~Chaminda Bandara and Vishal~M Patel.
\newblock A transformer-based siamese network for change detection.
\newblock {\em arXiv preprint arXiv:2201.01293}, 2022.

\bibitem[\protect\citeauthoryear{Chen and Shi}{2020a}]{levir}
Hao Chen and Zhenwei Shi.
\newblock A spatial-temporal attention-based method and a new dataset for
  remote sensing image change detection.
\newblock {\em Remote Sensing}, 12(10):1662, 2020.

\bibitem[\protect\citeauthoryear{Chen and Shi}{2020b}]{STANet}
Hao Chen and Zhenwei Shi.
\newblock A spatial-temporal attention-based method and a new dataset for
  remote sensing image change detection.
\newblock {\em Remote. Sens.}, 12:1662, 2020.

\bibitem[\protect\citeauthoryear{Chen \bgroup \em et al.\egroup
  }{2018}]{chen2018encoder}
Liang-Chieh Chen, Yukun Zhu, George Papandreou, Florian Schroff, and Hartwig
  Adam.
\newblock Encoder-decoder with atrous separable convolution for semantic image
  segmentation.
\newblock In {\em Proceedings of the European conference on computer vision
  (ECCV)}, pages 801--818, 2018.

\bibitem[\protect\citeauthoryear{Chen \bgroup \em et al.\egroup }{2022a}]{SaDL}
Hao Chen, Wenyuan Li, Song Chen, and Zhenwei Shi.
\newblock Semantic-aware dense representation learning for remote sensing image
  change detection.
\newblock {\em {IEEE} Transactions on Geoscience and Remote Sensing}, 60:1--18,
  2022.

\bibitem[\protect\citeauthoryear{Chen \bgroup \em et al.\egroup }{2022b}]{BIT}
Hao Chen, Zipeng Qi, and Zhenwei Shi.
\newblock Remote sensing image change detection with transformers.
\newblock {\em IEEE Transactions on Geoscience and Remote Sensing}, 60:3095166,
  2022.

\bibitem[\protect\citeauthoryear{Daudt \bgroup \em et al.\egroup }{2018a}]{FC}
Rodrigo~Caye Daudt, Bertr Le~Saux, and Alexandre Boulch.
\newblock Fully convolutional siamese networks for change detection.
\newblock In {\em 2018 25th IEEE International Conference on Image Processing
  (ICIP)}, pages 4063--4067. IEEE, 2018.

\bibitem[\protect\citeauthoryear{Daudt \bgroup \em et al.\egroup
  }{2018b}]{daudt2018urban}
Rodrigo~Caye Daudt, Bertr Le~Saux, Alexandre Boulch, and Yann Gousseau.
\newblock Urban change detection for multispectral earth observation using
  convolutional neural networks.
\newblock In {\em IGARSS 2018-2018 IEEE International Geoscience and Remote
  Sensing Symposium}, pages 2115--2118. Ieee, 2018.

\bibitem[\protect\citeauthoryear{Fang \bgroup \em et al.\egroup
  }{2021}]{SNUNet}
Sheng Fang, Kaiyu Li, Jinyuan Shao, and Zhe Li.
\newblock Snunet-cd: A densely connected siamese network for change detection
  of vhr images.
\newblock {\em IEEE Geoscience and Remote Sensing Letters}, 19:1--5, 2021.

\bibitem[\protect\citeauthoryear{He \bgroup \em et al.\egroup }{2016}]{resnet}
Kaiming He, Xiangyu Zhang, Shaoqing Ren, and Jian Sun.
\newblock Deep residual learning for image recognition.
\newblock In {\em Proceedings of the IEEE conference on computer vision and
  pattern recognition}, pages 770--778, 2016.

\bibitem[\protect\citeauthoryear{Hu \bgroup \em et al.\egroup
  }{2017}]{CA-SENet}
Jie Hu, Li~Shen, Samuel Albanie, Gang Sun, and Enhua Wu.
\newblock Squeeze-and-excitation networks.
\newblock {\em IEEE Transactions on Pattern Analysis and Machine Intelligence},
  42:2011--2023, 2017.

\bibitem[\protect\citeauthoryear{Hussain \bgroup \em et al.\egroup
  }{2013}]{hussain2013change}
Masroor Hussain, Dongmei Chen, Angela Cheng, Hui Wei, and David Stanley.
\newblock Change detection from remotely sensed images: From pixel-based to
  object-based approaches.
\newblock {\em ISPRS Journal of photogrammetry and remote sensing}, 80:91--106,
  2013.

\bibitem[\protect\citeauthoryear{Jaderberg \bgroup \em et al.\egroup
  }{2015}]{SA}
Max Jaderberg, Karen Simonyan, Andrew Zisserman, and Koray Kavukcuoglu.
\newblock Spatial transformer networks.
\newblock In {\em NIPS}, 2015.

\bibitem[\protect\citeauthoryear{Ji \bgroup \em et al.\egroup }{2018}]{whu}
Shunping Ji, Shiqing Wei, and Meng Lu.
\newblock Fully convolutional networks for multisource building extraction from
  an open aerial and satellite imagery data set.
\newblock {\em IEEE Transactions on Geoscience and Remote Sensing},
  57(1):574--586, 2018.

\bibitem[\protect\citeauthoryear{Ji \bgroup \em et al.\egroup
  }{2019}]{Two-stage3}
Shunping Ji, Yanyun Shen, Meng Lu, and Yongjun Zhang.
\newblock Building instance change detection from large-scale aerial images
  using convolutional neural networks and simulated samples.
\newblock {\em Remote. Sens.}, 11:1343, 2019.

\bibitem[\protect\citeauthoryear{Krizhevsky \bgroup \em et al.\egroup
  }{2017}]{alexnet}
Alex Krizhevsky, Ilya Sutskever, and Geoffrey~E Hinton.
\newblock Imagenet classification with deep convolutional neural networks.
\newblock {\em Communications of the ACM}, 60(6):84--90, 2017.

\bibitem[\protect\citeauthoryear{Li \bgroup \em et al.\egroup
  }{2021}]{li2021simple}
Pan Li, Da~Li, Wei Li, Shaogang Gong, Yanwei Fu, and Timothy~M Hospedales.
\newblock A simple feature augmentation for domain generalization.
\newblock In {\em Proceedings of the IEEE/CVF International Conference on
  Computer Vision}, pages 8886--8895, 2021.

\bibitem[\protect\citeauthoryear{Li \bgroup \em et al.\egroup
  }{2022}]{TransUNetCD}
Qingyang Li, Ruofei Zhong, Xin Du, and Yuying Du.
\newblock Transunetcd: A hybrid transformer network for change detection in
  optical remote-sensing images.
\newblock {\em IEEE Transactions on Geoscience and Remote Sensing}, 60:1--19,
  2022.

\bibitem[\protect\citeauthoryear{Liu \bgroup \em et al.\egroup
  }{2019a}]{Two-stage2}
Ruoyun Liu, Monika Kuffer, and Claudio Persello.
\newblock The temporal dynamics of slums employing a cnn-based change detection
  approach.
\newblock {\em Remote. Sens.}, 11:2844, 2019.

\bibitem[\protect\citeauthoryear{Liu \bgroup \em et al.\egroup
  }{2019b}]{DTCDSCN}
Yi~Liu, Chao Pang, Zongqian Zhan, Xiaomeng Zhang, and Xue Yang.
\newblock Building change detection for remote sensing images using a dual-task
  constrained deep siamese convolutional network model.
\newblock {\em IEEE Geoscience and Remote Sensing Letters}, 18:811--815, 2019.

\bibitem[\protect\citeauthoryear{Nemoto \bgroup \em et al.\egroup
  }{2017}]{Two-stage1}
Keisuke Nemoto, Ryuhei Hamaguchi, M.~Sato, Aito Fujita, Tomoyuki Imaizumi, and
  Shuhei Hikosaka.
\newblock Building change detection via a combination of cnns using only rgb
  aerial imageries.
\newblock In {\em Remote Sensing}, 2017.

\bibitem[\protect\citeauthoryear{Vaswani \bgroup \em et al.\egroup
  }{2017}]{vaswani2017attention}
Ashish Vaswani, Noam Shazeer, Niki Parmar, Jakob Uszkoreit, Llion Jones,
  Aidan~N Gomez, {\L}ukasz Kaiser, and Illia Polosukhin.
\newblock Attention is all you need.
\newblock {\em Advances in neural information processing systems}, 30, 2017.

\bibitem[\protect\citeauthoryear{Verma \bgroup \em et al.\egroup
  }{2019}]{manifoldmix}
Vikas Verma, Alex Lamb, Christopher Beckham, Amir Najafi, Ioannis Mitliagkas,
  David Lopez-Paz, and Yoshua Bengio.
\newblock Manifold mixup: Better representations by interpolating hidden
  states.
\newblock In {\em International Conference on Machine Learning}, pages
  6438--6447. PMLR, 2019.

\bibitem[\protect\citeauthoryear{Wang \bgroup \em et al.\egroup }{2019}]{isda}
Yulin Wang, Xuran Pan, Shiji Song, Hong Zhang, Gao Huang, and Cheng Wu.
\newblock Implicit semantic data augmentation for deep networks.
\newblock {\em Advances in Neural Information Processing Systems}, 32, 2019.

\bibitem[\protect\citeauthoryear{Xie \bgroup \em et al.\egroup
  }{2021}]{xie2021segformer}
Enze Xie, Wenhai Wang, Zhiding Yu, Anima Anandkumar, Jose~M Alvarez, and Ping
  Luo.
\newblock Segformer: Simple and efficient design for semantic segmentation with
  transformers.
\newblock In {\em Neural Information Processing Systems (NeurIPS)}, 2021.

\bibitem[\protect\citeauthoryear{Yang \bgroup \em et al.\egroup
  }{2022}]{yang2022masked}
Zhendong Yang, Zhe Li, Mingqi Shao, Dachuan Shi, Zehuan Yuan, and Chun Yuan.
\newblock Masked generative distillation.
\newblock {\em arXiv preprint arXiv:2205.01529}, 2022.

\bibitem[\protect\citeauthoryear{Zhang and Shi}{2020}]{CDWork}
Min Zhang and Wenzhong Shi.
\newblock A feature difference convolutional neural network-based change
  detection method.
\newblock {\em IEEE Transactions on Geoscience and Remote Sensing},
  58:7232--7246, 2020.

\bibitem[\protect\citeauthoryear{Zhang \bgroup \em et al.\egroup
  }{2020a}]{dsifn}
Chenxiao Zhang, Peng Yue, Deodato Tapete, Liangcun Jiang, Boyi Shangguan,
  Li~Huang, and Guangchao Liu.
\newblock A deeply supervised image fusion network for change detection in high
  resolution bi-temporal remote sensing images.
\newblock {\em ISPRS Journal of Photogrammetry and Remote Sensing},
  166:183--200, 2020.

\bibitem[\protect\citeauthoryear{Zhang \bgroup \em et al.\egroup
  }{2020b}]{IFNet}
Chenxiao Zhang, Peng Yue, Deodato Tapete, Liangcun Jiang, Boyi Shangguan,
  Li~Huang, and Guangchao Liu.
\newblock A deeply supervised image fusion network for change detection in high
  resolution bi-temporal remote sensing images.
\newblock {\em Isprs Journal of Photogrammetry and Remote Sensing},
  166:183--200, 2020.

\bibitem[\protect\citeauthoryear{Zheng \bgroup \em et al.\egroup
  }{2021}]{changestar}
Zhuo Zheng, Ailong Ma, Liangpei Zhang, and Yanfei Zhong.
\newblock Change is everywhere: Single-temporal supervised object change
  detection in remote sensing imagery.
\newblock In {\em Proceedings of the IEEE/CVF International Conference on
  Computer Vision}, pages 15193--15202, 2021.

\end{thebibliography}

\end{document}